\documentclass[letterpaper]{article} 
\usepackage{aaai25}  
\usepackage{times}  
\usepackage{helvet}  
\usepackage{courier}  
\usepackage[hyphens]{url}  
\usepackage{graphicx} 
\usepackage{natbib}  
\usepackage{caption} 
\usepackage{algorithm}
\usepackage{algorithmic}
\usepackage{amsmath}
\usepackage{amsfonts}
\usepackage{booktabs}
\usepackage{multirow} 

\usepackage{color}
\usepackage{newfloat}
\usepackage{listings}
\usepackage{xcolor} 
\usepackage{color, colortbl}
\definecolor{mygray}{gray}{.9}

\newcommand{\answerYes}[1]{\textcolor{blue}{[Yes]}}
\newcommand{\answerPartial}[1]{\textcolor{blue}{[Partial]}}
\newcommand{\answerNo}[1]{\textcolor{red}{[No]}}
\newcommand{\answerNA}[1]{\textcolor{gray}{[NA]}}

\usepackage{newfloat}
\usepackage{listings}
\DeclareCaptionStyle{ruled}{labelfont=normalfont,labelsep=colon,strut=off} 
\floatstyle{ruled}
\newfloat{listing}{tb}{lst}{}
\floatname{listing}{Listing}
\title{pFedGPA: Diffusion-based Generative Parameter Aggregation for\\ Personalized Federated Learning}
\author{
    Jiahao Lai\textsuperscript{\rm 1},
    Jiaqi Li\textsuperscript{\rm 1},
    Jian Xu\textsuperscript{\rm 1},
    Yanru Wu\textsuperscript{\rm 1},
    Boshi Tang\textsuperscript{\rm 1},
    Siqi Chen\textsuperscript{\rm 1},\\
    Yongfeng Huang\textsuperscript{\rm 2},
    Wenbo Ding\textsuperscript{\rm 1},
    Yang Li\textsuperscript{\rm 1}\footnote{Corresponding author.}
}

\affiliations{
    \textsuperscript{\rm 1} Tsinghua Shenzhen International Graduate School, Tsinghua University\thanks{Jiahao Lai, Jian Xu, Yanru Wu, Siqi Chen, Wenbo Ding and Yang Li are from the Shenzhen Key Laboratory of Ubiquitous Data Enabling, Tsinghua University.}\\
    \textsuperscript{\rm 2} Department of Computer Science and Engineering, The Chinese University of Hong Kong\\
    laijh22@mails.tsinghua.edu.cn,
    yangli@sz.tsinghua.edu.cn
}

\usepackage{bibentry}

\begin{document}

\maketitle

\begin{abstract}
Federated Learning (FL) offers a decentralized approach to model training, where data remains local and only model parameters are shared between the clients and the central server. Traditional methods, such as Federated Averaging (FedAvg), linearly aggregate these parameters which are usually trained on heterogeneous data distributions, potentially overlooking the complex, high-dimensional nature of the parameter space. This can result in degraded performance of the aggregated model. While personalized FL approaches can mitigate the heterogeneous data issue to some extent, the limitation of linear aggregation remains unresolved. To alleviate this issue, we investigate the generative approach of diffusion model and propose a novel generative parameter aggregation framework for personalized FL, \texttt{pFedGPA}. In this framework, we deploy a diffusion model on the server to integrate the diverse parameter distributions and propose a parameter inversion method to efficiently generate a set of personalized parameters for each client. This inversion method transforms the uploaded parameters into a latent code, which is then aggregated through denoising sampling to produce the final personalized parameters. By encoding the dependence of a client's model parameters on the specific data distribution using the high-capacity diffusion model, \texttt{pFedGPA} can effectively decouple the complexity of the overall distribution of all clients' model parameters from the complexity of each individual client's parameter distribution. Our experimental results consistently demonstrate the superior performance of the proposed method across multiple datasets, surpassing baseline approaches.
\end{abstract}

%

\begin{figure}[t]
	\centering
	\includegraphics[width=0.45\textwidth]{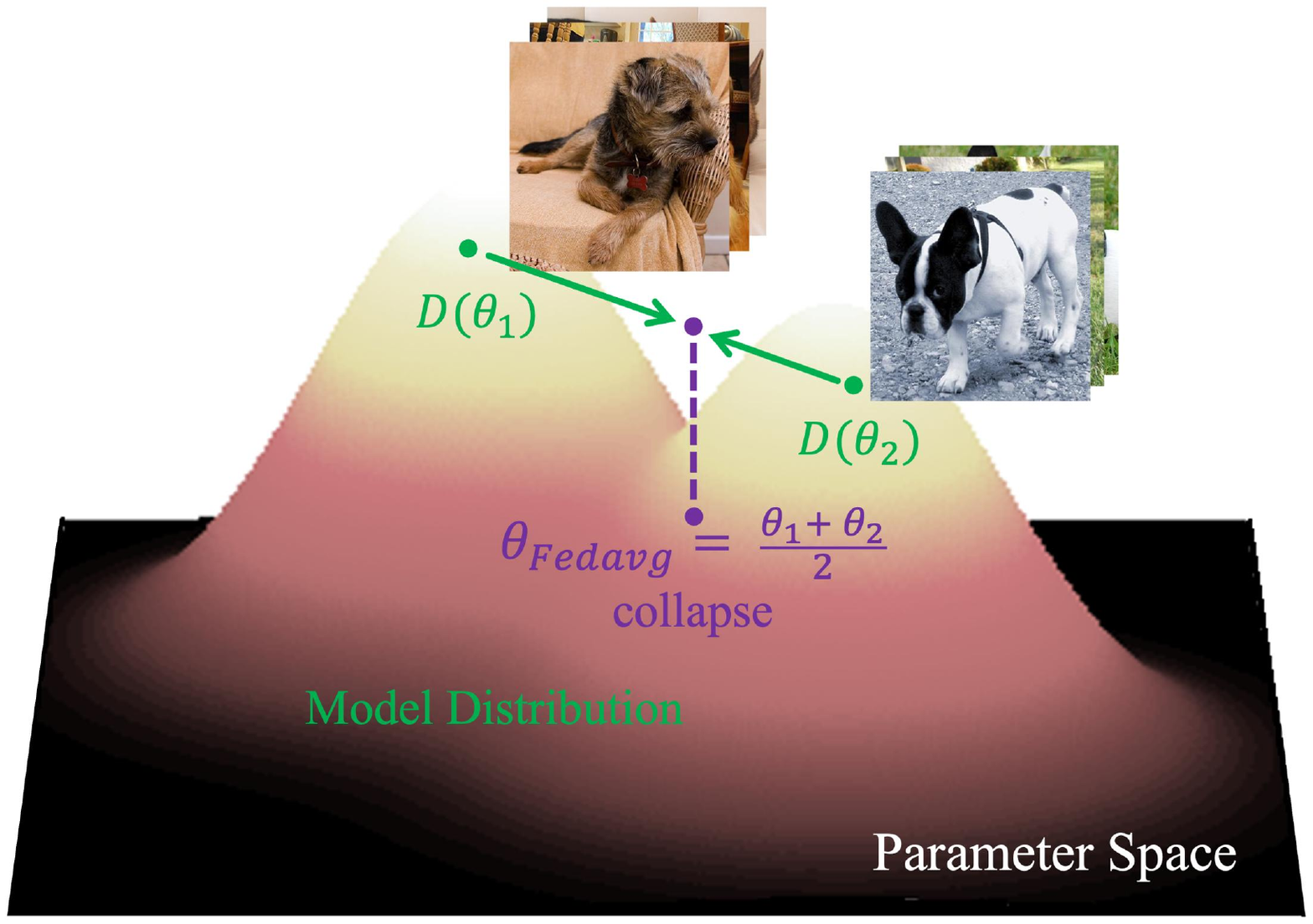}
	\caption{Parameter collapse can occur when linearly averaging the parameters from different clients. Bright colors indicate high-probability regions of the parameter space, where the parameters located at the peaks of the model distribution are well-optimized for specific tasks.}
	\vspace{-17pt}
	\label{mode_collapse}
\end{figure}

\section{Introduction}
To meet with the increasing needs of privacy protection, federated learning (FL) has been a popular machine learning paradigm and research topic for many years \cite{mcmahan2017communication}. Typically, multiple devices such as smartphones, sensors, and IoTs (Internet of Things) collaboratively train a global model under the coordination of a central server. 
FL has been extended to various application domains, including healthcare and finance \cite{nguyen2022health}. In healthcare, FL trains models on distributed patient data, enabling cooperative research without compromising privacy \cite{yang2019federated}. In finance, FL helps developing fraud detection models using data from multiple financial institutions while ensuring confidentiality \cite{li2019privacy}.

One challenge in FL is handling the heterogeneity of data distributions across various devices \cite{zhao2018federated,kairouz2021advances}. For example, in healthcare applications, patient data collected from different hospitals can vary greatly due to differences in patient demographics, medical equipment, and local practices. In linear aggregation methods such as FedAvg \cite{mcmahan2017communication}, data in different clients are assumed to share the same distribution. Non-IID data in local clients can lead to unstable results \cite{zhao2018federated,Karimireddy20SCAFFOLD}, as illustrated in Fig. ~\ref{mode_collapse}.
In this case, the parameters $\theta_1$ and $\theta_2$ from non-IID datasets are aggregated to form $\theta_{Fedavg} = (\theta_1 + \theta_2)/2$. While $\theta_1$ and $\theta_2$ are located in the high-probability regions of the parameter space and are well-optimized for their respective tasks, their average $\theta_{Fedavg}$ falls into a low-probability region, causing a collapse. This example highlights the limitations of linear aggregators and the need of more ingenious aggregation methods.

To handle Non-IID client data, various personalized FL methods have been proposed. They aim to learn a customized model for each individual client to fit its local data distribution. Although many efforts have been devoted to develop advanced model adaptation methods to obtain better local models, the limitation of linear aggregation remains since most of the advanced personalized FL methods still highly rely on the linear aggregation \cite{Collins21FedRep,sattler2020clustered,Yang23FedCap,xu2023personalized}. For instance, a simple but tough-to-beat baseline is combining FedAvg with local fine-tuning, where the quality of the global model by FedAvg may directly affect the resulted local models.

To build better FL models, a fundamental challenge is to explicitly model the relationship between data distributions and model parameters within specific optimization constraints. However, when considering the parameter space where the model distribution resides, we recognize that the intrinsic difficulty arises from modeling this distribution as a low-dimensional manifold within a high-dimensional space \cite{zhou2022towards, zhou2023understanding}. Simple network architectures like multi-layer perceptrons (MLPs) or basic Transformers struggle to effectively learn such complex distributions and scale robustly. Moreover, they lack effective optimization objectives and training algorithms.

Inspired by the remarkable success of diffusion models in achieving state-of-the-art results in image generation \cite{dhariwal2021diffusion} and recent attempts in model parameter generation \cite{wang2024neural}, we propose to leverage the diffusion models to capture the distribution of model parameters in the parameter space. Diffusion models, which convert the complex data distribution to an analytically tractable prior, estimate the score function at each time step and iteratively refine samples through stochastic processes \cite{song2020score}. Although initially applied to image data, this versatile approach shows promise for modeling high-dimensional parameter distributions \cite{zhang2024metadiff,wang2024neural}.

Building on above idea, we introduce our framework, \texttt{pFedGPA}, which employs a diffusion model on the server to handle the parameters of clients, enabling the server to learn the distributions of all clients' model parameters using a powerful generative model. Subsequently, we investigate the mechanisms of generative parameter aggregation and personalized model updates within this framework. We consider the generation process as an alternative to achieve the better model aggregation. Our method effectively guides the optimization direction for newly joined clients and significantly accelerates the initialization. Furthermore, we develop a novel parameter inversion method inspired by image editing techniques applied within diffusion models, which transforms the original parameters into a latent representation and then refines them to generate new parameters that retain the implicit semantics of the original data while incorporating the learned patterns from the diverse parameter distributions. Furthermore, we explore the configuration of the diffusion model architecture tailored for processing model parameter data. Our experimental results demonstrate the effectiveness of the proposed method, consistently achieving superior performance across multiple datasets compared to baseline approaches. The main contributions of our work are concluded as follows:
\begin{itemize}
	\item We present a novel FL framework that employs diffusion models to generate personalized parameters with the global guidance in heterogeneous settings. To the best of our knowledge, we are the first to improve model aggregation by applying diffusion models in FL.
	\item We conducted experiments to evaluate the generation quality by producing partial and full parameters for small models, and partial parameters for large models.
	\item We empirically verify the superiority of our proposed method in terms of generalization performance on several benchmark datasets under practical scenarios.
\end{itemize}

\section{Related Work} \label{sec:related-work}
\subsection{FL with Data Heterogeneity}
To address the non-IID data challenge, several strategies for improving the global models have been proposed. FedProx \cite{li2020federated} introduced a local regularization term to optimize each client's model. SCAFFOLD \cite{karimireddy2020scaffold} introduces control variates to correct the drift in local updates. Clustering-based methods \cite{sattler2020clustered, li2020federated} cluster similar local models into groups and assign each group a global model. Moreover, self-supervised methods have been incorporated to define the similarity between model representations for correcting local training \cite{li2021model}. Personalized FL \cite{smith2017federated} focuses on tailoring models for individual clients by combining information across clients. Decentralized MAML \cite{fallah2020personalized} adapts the model-agnostic meta-learning framework to a federated setting, allowing clients to learn models through local adaptation \cite{finn2017model, nichol2018first}. Another category includes model mixing and layer adaptation strategies, where clients learn a mixture of global and local models \cite{hanzely2020federated,zhang2023fedala}. Decoupling model layers is also a popular and effective approach \cite{arivazhagan2019fedper,Collins21FedRep,Xu22FedPer++}. For instance, FedRep \cite{Collins21FedRep} and FedBABU \cite{OhKY22FedBABU} train base layers globally using FedAvg, with personalized layers fine-tuned locally. Personalized aggregation based on model similarity has also been investigated \cite{Zhang21FedFOMO,ye23fedgraph}.

\subsection{Diffusion Models}
Diffusion models have emerged as a powerful technique of generative AI, particularly excelling in producing high-quality images. Their superior properties have led to widespread applications in various vision \cite{lugmayr2022repaint} and multi-modal tasks \cite{rombach2022high, kawar2023imagic, kumari2023multi}. The foundational work by \cite{sohl2015deep}, studied non-equilibrium thermodynamics, highlighting its potential in generative modeling. \cite{ho2020denoising} advanced this with Denoising Diffusion Probabilistic Models (DDPMs), improving the denoising process for better image synthesis. \cite{song2019generative} proposed Score-Based Generative Modeling, achieving remarkable results in diverse applications \cite{meng2021sdedit, xu2022geodiff}. 
\cite{song2020score} further unified diffusion probabilistic modeling and score-based generative modeling using stochastic differential equations (SDEs), demonstrating their equivalence.

Guided diffusion models \cite{dhariwal2021diffusion} introduced mechanisms to steer the generative process. Notable examples include DALL·E 2 \cite{ramesh2022hierarchical}. In contrast, classifier-free guidance \cite{ho2022classifier} integrates guidance directly into the model by conditioning on input prompts, thus avoiding the need for a separate classifier. Examples of this approach include Imagen \cite{saharia2022photorealistic} and Stable Diffusion \cite{rombach2022high}.

\subsection{Model Parameter Generation}
Model parameter generation has progressed from gradient optimization to advanced meta-learning and Bayesian methods. Early efforts primarily focused on optimizing parameters using gradient-based techniques like Stochastic Gradient Descent (SGD) and its variants \cite{amari1993backpropagation}. Meta-learning emerged to enable models to adapt quickly to new tasks with minimal data, exemplified by methods that use hypernetworks as metaknowledge to dynamically generate model parameters from input data \cite{zhmoginov2022hypertransformer}. Bayesian deep learnig used Variational Bayes to infer the distributions of model parameters, which are assumed to be Gaussian \cite{wilson2020bayesian}.

Recently, diffusion models have emerged as a powerful paradigm for model parameter generation. G.pt \cite{peebles2022learning} trains conditional diffusion models using model checkpoints as datasets. MetaDiff \cite{zhang2024metadiff} introduces a diffusion-based meta-learning method for few-shot tasks. HyperDiffusion \cite{erkocc2023hyperdiffusion} uses diffusion models to generate new neural implicit fields. Additionally, p-diff \cite{wang2024neural} examines the quality of model parameters generated through diffusion. These works provide preliminary experiments and analyses on using diffusion models for model parameter generation.

\section{Problem Definition and Preliminaries} \label{sec:problem}
\subsection{Federated Learning Setting}
FL aims to collectively train a centralized model for $n$ edge-distributed clients. Each client $i$ has access to $m_i$ data samples $D_i={(x_j^{(i)}, y_j^{(i)})}_{j=1}^{m_i}$ from its own private data distribution $P_i$ on $\mathcal{X}\times\mathcal{Y}$ and $N$ is the total number of data samples overall clients. Generally, the data distributions for each client are different, i.e., $P_i \ne P_j$ for any pair $i, j \in {1, \dots, n}$.
Let $\ell_i: \mathcal{Y} \times \mathcal{Y} \to \mathcal{R}_+$ denote the loss function corresponding to client $i$, and $F_i: \Theta \times \mathcal{X} \to \mathcal{Y}$ denote the local model parameterized by $\theta^i$. The goal of conventional FL is to optimize the following objectives:
\begin{equation}
\theta^* = \arg\min_{\theta} \frac{1}{n} \sum_{i=1}^n \mathbb{E}_{(x,y) \sim P_i}[\ell_i(F_i(\theta, x), y)],
\label{eq:conventional_FL}
\end{equation}
where $\theta^*$ represents the globally optimal parameters. The representative method for solving Eq.~(\ref{eq:conventional_FL}) is FedAvg \cite{mcmahan2016communication}. In each round, clients perform several epochs of SGD on their local loss functions and send the updated models to the server. The server then aggregates the models from all clients linearly by
\(  \bar{\theta} = \sum_{i=1}^{n} \frac{m_i}{N}\theta^i \) and broadcasts the averaged model $\bar{\theta}$ back to the clients.

Moreover, due to data heterogeneity, the unified parameters may not be locally optimal for each client. Therefore, personalized FL adjusts the optimization objective to:
\begin{equation}
\Theta^* = \arg\min_{\Theta:=\{\theta^i\}_{i=1}^n} \frac{1}{n} \sum_{i=1}^n \mathbb{E}_{(x,y) \sim P_i}[\ell_i(F_i(\theta^i; x), y)].
\label{eq:PFL}
\end{equation}
where $\Theta^*$ denotes the collection of locally optimal parameters. The challenge lies in how to aggregate model parameters from heterogeneous clients on the server and produce parameters that incorporate insights from all clients while still adapting to specific data distributions. In this work, we propose to train a diffusion model at the server to address these challenges.

\subsection{Diffusion Probabilistic Models}
Here, we focus on the Denoising Diffusion Probabilistic Models formulation, primarily because of its prevalence and consistency with the Denoising Score Matching using Langevin Dynamics. DDPMs gradually add noise to data, transforming it into standard Gaussian noise, and then learn to denoise step-by-step, generating new data. It involves a forward process and a reverse process as described below.

\paragraph{Forward process.}  Given training data $z_0$ from a target
distribution $q(z_0)$, the forward process gradually adds 
Gaussian noise to diffuse $z_0$ into $z_1, z_2,\dots, z_T$, where $z_T$ is approximately sampled from the standard Gaussian distribution $q(z_{T}) \approx  \mathcal{N}(z_{T};\mathbf{0}, \mathbf{I})$. This process can be formulated as follows:

\begin{equation}
q(z_{1:T}|z_{0}) = \prod_{t = 1}^{T} q(z_{t}| z_{t-1})
\label{eq:dm1}
\end{equation}

\begin{equation}
q(z_{t}|z_{t-1}) = \mathcal{N}(z_{t};\sqrt{1 - \beta_t}z_{t-1}, \beta_t\mathbf{I})
\label{eq:dm2}
\end{equation}
where $\beta_1, \dots, \beta_T \in (0,1)$ is a variance schedule. The diffusion model aims to approximate the noise added at each time step $t$ using a neural estimator $\epsilon_{\phi  }(z_{t},t)$. The training objective $L_{ddpm}$ is given by the Eq.~(\ref{eq:L_ddpm}) below:
\begin{equation}
	\begin{aligned}
        L_{ddpm}=\mathbb{E}_{t \sim [1,T], z_{0} \sim q_0, \epsilon_t \sim \mathcal{N}(\mathbf{0}, \mathbf{I})} [\|\epsilon_t-\epsilon_{\phi}(z_{t}, t)\|_2^2],
	\end{aligned}
	\label{eq:L_ddpm}
\end{equation}
where $z_{t} = \sqrt{\bar{\alpha}_{t} } z_{0} + \sqrt{1- \bar{\alpha}_{t} }\epsilon_{t}$ and $\bar{\alpha}_{t} =  {\textstyle \prod_{j=1}^{t}} (1 - \beta_j)$. 

\paragraph{Reverse process.} During the reverse process, we iteratively reconstruct the target data $z_{0}$ from the random noise $z_T \sim \mathcal{N}(\mathbf{0}, \mathbf{I})$. At each time step $t$, we sample $z_{t-1}$ from the estimated reverse Markov chain $p_{\phi}(z_{t}| z_{t-1})$ parameterized by the noise estimator $\epsilon_{\phi}(z_{t}, t)$. That is, 
\begin{equation}
	\begin{aligned}
		z_{t-1}=\frac{1}{\sqrt{\alpha_{t}}}(z_{t}-\frac{\beta_t}{\sqrt{(1-\overline{\alpha}_{t})}}\epsilon_{\phi}(z_{t}, t)) + \sigma_{t}z_t.
	\end{aligned}
	\label{eq:denoising_process}
\end{equation}
where $z_t \sim \mathcal{N}(\mathbf{0}, \mathbf{I})$ and $\sigma_{t}^2 = \frac{1-\bar{\alpha}_{t-1}}{1-\bar{\alpha}_{t}}\beta_t$ represents the derived variance. 

\paragraph{Analysis.} As demonstrated in \cite{song2020score}, DDPMs can be viewed as discretizations to SDEs, coherently bridging diffusion probabilistic modeling and noise conditional score networks into a unified framework. Thus,
we can reformulate $L_{ddpm}$ equivalently as denoising score matching:
\begin{equation}
	\begin{aligned}
        \mathbb{E}_{q(z_0)}\mathbb{E}_{q(z_t|z_0)} [\|s_{\phi}(z_t,t)-\nabla_{z_t}\log_{}{q(z_t|z_0)} \|_2^2], 
	\end{aligned}
	\label{eq:L_ddpm1}
\end{equation}
where $s_{\phi}(z_t,t) = -\frac{\epsilon_{\phi}(z_{t}, t)}{\sqrt{(1-\overline{\alpha}_{t})}}$. This allows us to use DDPMs to effectively estimate the scores of data distributions.

\section{Methodology} \label{sec:methodology}
In this section, we present our framework \texttt{pFedGPA} in detail. Given a collection of clients with their own specific data distributions in a FL system, our goal is to integrate the diverse parameter distributions captured by the diffusion model on the server, which reflects the underlying data distributions of the clients. Meanwhile, the diffusion model aims to generate new, personalized model parameters for each client by leveraging these learned parameter distributions. In the following subsections, we first illustrate the parameter aggregation using the diffusion model. Next, we explain the personalized parameter generation. Finally, we delve into the architecture design of the diffusion model. More details about the training procedure are presented in Algorithm~\ref{alg:fed_diffusion}.

\begin{figure}[t!]
	\centering
	\includegraphics[width=0.48\textwidth]{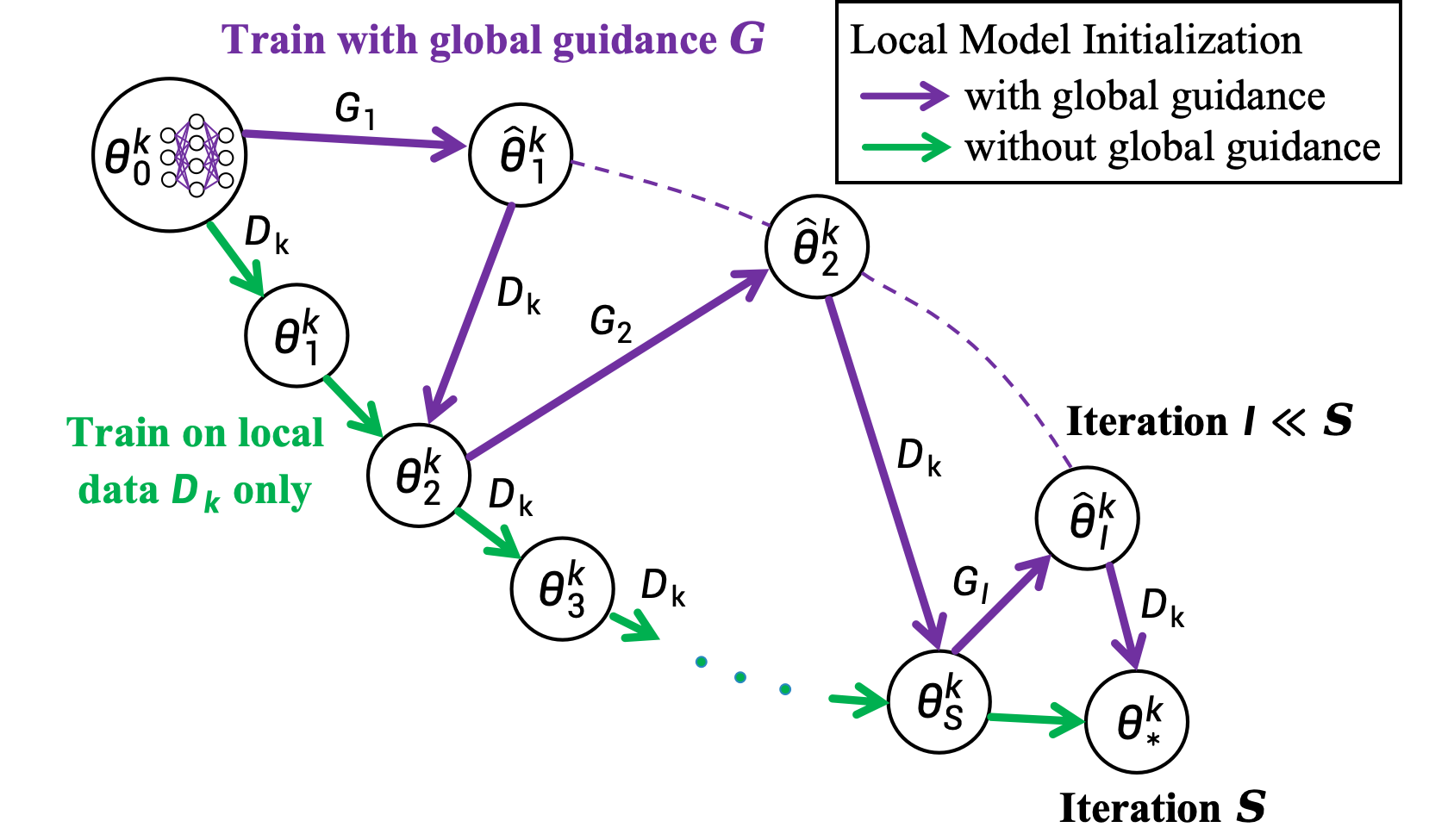}
        \vspace{-3ex}
	\caption{
Illustration of the training process for a new client 
$k$ with and without global guidance. The green arrows represent training starting from the initial parameters  $\theta^{k}_{0}$ solely on local data $D_{k}$, gradually converging to the final optimized parameters $\theta^{k}_{*} $ within $S$ iterations. The purple arrows indicate training with global guidance $G$ alternated with local data training, which accelerates initialization and converges within $I \ll S$ iterations.
}
	\vspace{-5pt}
	\label{fast_adaptation}
\end{figure}

\subsection{Generative Parameter Aggregation}
In our framework, we deploy a diffusion model on the server, where both input and output are the weights of the clients' models. The core of this approach is to enable the server to learn the distribution of all clients' model parameters using a powerful generative model, a distribution which resides on a low-dimensional manifold in the parameter space.

Let $ H: \Phi  \times \Theta   \to \Theta$ denotes the diffusion model parametrized by $\phi$ and we modify the optimization objective in Eq.~(\ref{eq:PFL}) based on the parameter generation to derive:
\begin{equation}
\phi^*,\Theta^* = \arg\min_{\phi,\Theta:=\{\theta^i\}_{i=1}^n} \frac{1}{n} \sum_{i=1}^n \mathcal{L}_i(\phi,\theta^i),
\label{eq:diffusion_PFL}
\end{equation}
where $\mathcal{L}_i(\phi,\theta^i) = \mathbb{E}_{(x,y) \sim P_i}[\ell_i(F_i(H(\phi;\theta^i); x), y)]$ and $\{\theta^i_*\}_{i=1}^n$ are the stationary points of the system $H$ under well-fitted state $\phi^*$ that represent the optimal model parameters of each client. In each round, the server receives the client parameters and treats them as training data for the diffusion model, training it over a predefined number of epochs. By performing parameter inversion, which will be introduced in the next section, the server then generates personalized model parameters for each client.

\paragraph{Global Guidance.} To quickly initialize a newly joined client, we propose a global guidance approach to help the client model adapt its parameters to the local data in just a few iterations.
In \cite{shamsian2021personalized}, a hypernetwork is used to take a trainable vector descriptor as input to differentiate each client. However, this approach has scalability limitations. Specifically, when new clients join the network, it often requires training new embedding vectors from scratch, which becomes increasingly inefficient as the network grows. Moreover, fine-tuning all clients' embedding vectors to achieve consistent representation is necessary but complex.
In diffusion models, if the input also includes client descriptors $e_{client}$, it can be implemented as a form of classifier-free guidance, utilizing a linear combination of conditional and unconditional score estimates to provide client-specific guidance:
\begin{equation}
\tilde{\epsilon}_{\phi}(\theta, e_{client}) = (1 + \omega )\epsilon_{\phi}(\theta, e_{client}) - \omega \epsilon_{\phi}(\theta),
\end{equation}
where $\omega$ controls the strength of the guidance. The advantage is that there is no need to train a separate classifier. However, in our approach, we adopt classifier-based guidance because, in the FL setting, each client inherently serves as a classifier relative to the diffusion model. In turn, the diffusion model offers global guidance to the local client models. That is:
\begin{equation}
\tilde{\epsilon}_{\phi}(\theta) = \underbrace{\epsilon_{\phi}(\theta)}_{global\;  guidance} - \underbrace{(1 + \omega )\nabla_{\theta} \log P_{y}(x| \theta)}_{local\; update},
\end{equation}
where $\log P_{y}(x| \theta)$ represents the training loss of the client model in probabilistic form (e.g., cross-entropy loss). This setup allows the diffusion model to share information across clients while preserving the adaptability of personalized models. Instead of fitting each client's distribution individually, a unified model is used to estimate the overall score of model distributions, thereby facilitating efficient knowledge sharing.

For newly joined clients, this approach enables rapid initialization within just a few iterations. In each iteration, the client sends its fine-tuned model parameters to the server, where the diffusion model provides global guidance to update these parameters. This process is illustrated in Fig. \ref{fast_adaptation}.

\begin{figure}[t!]
	\centering
	\includegraphics[width=0.3\textwidth]{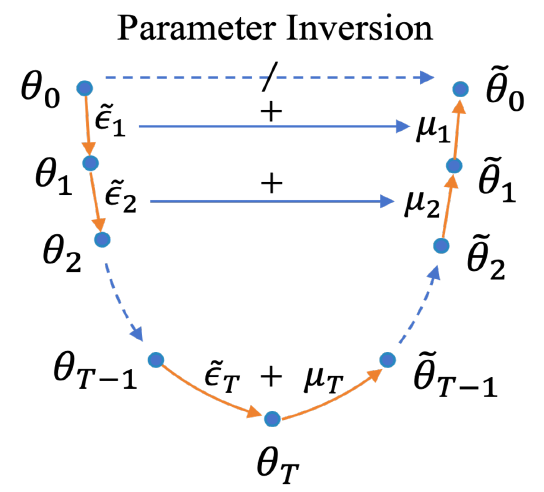}
        \vspace{-1ex}
	\caption{
Illustration of the Parameter Inversion. Starting with the initial parameter $\theta_{0}$, it is diffused through several steps to reach $\theta_{1}, \dots, \theta_{T}$. During this process, the noise introduced between consecutive time steps and the final state $\theta_{T}$ are recorded as the latent code for $\theta_{0}$. In the denoising sampling phase of the diffusion model, these elements are gradually encoded to produce a new parameter $\tilde{\theta}_{0}$. Notice that it is challenging to obtain $\tilde{\theta}_{0}$ directly from $\theta_{0}$ using a linear aggregator.
}
	\vspace{-4pt}
	\label{inversion}
\end{figure}

\subsection{Parameter Inversion}
Generating personalized parameters remains challenging because a single diffusion model cannot directly control outputs for individual clients. In image generation, control is often achieved through conditioning on labels, but in our scenario, no such labels are available. To address this, we propose a novel parameter inversion method inspired by unsupervised inversion techniques \cite{wu2023latent}. In our approach, the uploaded model parameters serve as implicit 'labels', which are decomposed into latent codes and injected into the diffusion model to generate stable, personalized parameters for each client.

Detailed explanation is as follows. First, in the forward process of the diffusion algorithm, Gaussian noises are added at each time step, diffusing $\theta_{0}$ to $\theta_{1},\dots,\theta_{T}$.
Next, we concatenate all the added noises with the final $\theta_{T}$ to define a latent code of $\theta_{0}$. The formulations are as follows:
\begin{gather}
\theta_{1},\dots,\theta_{T} \sim p(\theta_{1:T}|\theta_{0}), \\
\tilde{\epsilon_{t}} = (\theta_{t} - \sqrt{1 - \beta_t}\theta_{t-1} ) / \sqrt{\beta_t}, \,\,\,  t = T,\dots ,1, \\
\mathbf{\gamma} : = (\theta_{T} \oplus \tilde{\epsilon}_{T} \oplus \cdots \oplus \tilde{\epsilon}_{2} \oplus \tilde{\epsilon }_{1}),
\end{gather}
where the latent code $\mathbf{\gamma}$ encodes the implicit semantics of $\theta_{0}$. We can perfectly reconstruct $\theta_{0}$ from $\mathbf{\gamma}$: 
\begin{equation}
\theta_{T}\overset{\tilde{\epsilon}_{T}}{\rightarrow} \theta_{T-1}\overset{\tilde{\epsilon}_{T-1}}{\rightarrow}\cdots \overset{\tilde{\epsilon}_{1}}{\rightarrow} \theta_{0},
\end{equation}
here $\theta_{t} \overset{\tilde{\epsilon}_{t}}{\rightarrow} \theta_{t-1}$ means $\theta_{t-1} = (\theta_{t} - \sqrt{\beta_t}\tilde{\epsilon}_{t}) / \sqrt{1-\beta_t}$.
Furthermore, we aim to encode the latent code $\mathbf{\gamma}$ into the generation process of the new parameter $\tilde{\theta}_{0}$. To achieve this, during the denoising process of the diffusion model, we start sampling from $\theta_T$, and at each time step, we substitute the randomly sampled Gaussian noise $z_t$ in Eq.~(\ref{eq:denoising_process}) with the deterministic $\tilde{\epsilon}_{t}$ to inject the implicit semantics of $\theta_0$:
\begin{equation}
\tilde{\theta}_{t-1} = \underbrace{\mu_{\phi }(\tilde{\theta}_{t},t)}_{denoising \ direction} - \underbrace{\sigma_{t}\tilde{\epsilon}_{t}}_{implicit \ semantics},
\label{eq:inversion}
\end{equation}
where $\mu_{\phi }(\tilde{\theta}_{t},t)$ is the mean estimator in Eq.~(\ref{eq:denoising_process}):
\begin{equation}
\mu_{\phi }(\tilde{\theta}_{t},t) = \frac{1}{\sqrt{\alpha_{t}}}(\tilde{\theta}_{t}-\frac{\beta_t}{\sqrt{(1-\overline{\alpha}_{t})}}\epsilon_{\phi}(\tilde{\theta}_{t}, t)).
\end{equation}
This process is depicted in Fig. ~\ref{inversion}, which illustrates how the generated parameters retain the implicit semantics of the original parameters while effectively incorporating global information. This approach ultimately enhances the generalization performance of the model.

\begin{algorithm}[t]
	\caption{pFedGPA} 
        \label{alg:fed_diffusion}
	\hspace*{0.02in} {\bf Input:} 
	Communication rounds $T$, initialization rounds $I$\\
	\hspace*{0.02in} {\bf Server executes:}
	\begin{algorithmic}[1]
		\FOR{each round $t = 1,2,...,T$} 
            \IF{client $i$ uploads its model $\theta^{i}$}
               \STATE Update diffusion model with $\theta^{i}$ according to the loss $L_{ddpm}$
               \STATE Update local model $\hat{\theta }_{i}$ using Inversion by Eq.~(\ref{eq:inversion})
               \STATE $\theta^{i} \leftarrow$ LocalUpdate $(i, \hat{\theta }^{i})$
            \ENDIF
            \IF{new client $k$ joins the network}
                \STATE Initialize local model as $\hat{\theta}^k_0 = \theta^k_0$ 
                \FOR{each round $l = 1,2,...,I$}
                \STATE $\theta^k_l \leftarrow$ LocalUpdate $(k, \hat{\theta}^k_{l-1})$
                \STATE $\tilde{\epsilon}_{\phi}(\theta^k_l) = \epsilon_{\phi}(\theta^k_l) - (1 + \omega )(\theta^k_{l-1} - \theta^k_{l})$
                \STATE Update local model $\hat{\theta }^{k}_l$ using denoising sampling by Eq.~(\ref{eq:denoising_process}) iteratively for $s$ steps          
		    \ENDFOR
                \STATE $\theta^k \leftarrow$ LocalUpdate $(k, \hat{\theta}^k_{I})$
            \ENDIF
		\ENDFOR        
	\end{algorithmic} 
	\hspace*{0.02in} \\
	\hspace*{0.02in} {\bf LocalUpdate}$(i, \hat{\theta } ^{i})$:
	\begin{algorithmic}[1]
             \STATE Update local model: $\theta^{i} \gets \hat{\theta } ^{i}$.
		\FOR{each batch $\left(x, y\right) \in D_i$}
		\STATE Update local model: $\theta^{i} \gets \theta ^{i} - \lambda \nabla_{\theta}\ell_i(F_i(\theta^i; x), y)$
		\ENDFOR
		\RETURN $\theta^{i}$
	\end{algorithmic}
	\label{alg1}
\end{algorithm}

\begin{table*}[t]
    \centering
    \normalsize
    \resizebox{\textwidth}{!}{ 
    \begin{tabular}{ l c c c c c c c c c}
        \toprule[1pt]
        \multirow{2}{*}{Method}& \multicolumn{3}{c}{EMNIST}&\multicolumn{3}{c}{Fashion-MNIST} &
        \multicolumn{3}{c}{CIFAR-10}\\
        \cmidrule(lr){2-4}\cmidrule(lr){5-7}\cmidrule(lr){8-10}
        & 10 clients & 20 clients & 100 clients
        & 10 clients & 20 clients & 100 clients
        & 10 clients & 20 clients & 100 clients \\
        \midrule
        Local-only
        & 70.85 & 70.45 & 73.73
        & 85.11 & 84.89 & 85.48
        & 63.83 & 64.38 & 65.41
        \\
        FedAvg
        & 71.10 & 72.89 & 74.60
        & 81.90 & 82.96 & 83.93
        & 64.13 & 68.51 & 69.87 
        \\
        FedAvg-FT
        & 81.91 & \textbf{85.09} & 86.52
        & 88.45 & \textbf{89.39} &  89.28
        & 71.76 & 76.46 & 76.55 
        \\
        FedPer
        & 74.56 & 76.61 & 77.08
        & 87.13 & 87.20 & 88.65
        & 63.20 & 63.68 & 69.43
        \\
        FedRep
        & 74.22 & 74.37 & 76.54
        & 88.69 & 88.27 & 88.42
        & 64.71 & 65.85 & 67.63
        \\
        LG-FedAvg
        & 71.23 & 70.90 & 76.25
        & 85.28 & 85.25 & 85.96
        & 64.20 & 64.94 & 65.41
        \\
        FedBABU
        & 80.28 & 83.92 & 84.39
        & 87.44 & 88.86 & 89.43
        & 69.42 & 74.25 &75.32
        \\
        pFedHN
        & 77.33 & 81.79 & 77.23
        & 87.69 & 86.84 & 86.25
        & 71.43 & 75.41 & 77.12
        \\
        \midrule
        \textbf{pFedGPA}
        & \textbf{82.53} & \underline{84.70} & \textbf{88.22}
        & \textbf{88.90} & \underline{89.18} & \textbf{90.08}
        & \textbf{72.95} & \textbf{78.54} & \textbf{78.33}
        \\
        \bottomrule[1pt]
    \end{tabular}}
    \caption{The comparison of final round average test accuracy (\%) across different datasets: full participation with 10 and 20 clients, and client sampling at 30\% with 100 clients in the FL system. The best performance is in \textbf{bold} font and the second best is marked with \underline{underline}.}
    \label{tab:result}
\end{table*}

\subsection{Diffusion Model Designs}
We use the latent diffusion model \cite{rombach2022high} as the generative model and adopt the main architecture from p-diff \cite{wang2024neural}. The latent diffusion model consists of an autoencoder and a diffusion model. The autoencoder is first trained using reconstruction loss to produce a low-dimensional latent space. Then, the diffusion model operates within this latent space, learning to progressively denoise samples from noise, thus approximating the distribution of the latent representations. This approach enables the diffusion model to generate high-quality samples with reduced computational costs and memory usage compared to operating directly in the original high-dimensional space.

We flatten the model parameters into one-dimensional vectors and use 1-D convolution layers within both the encoder and decoder of the autoencoder, as well as the diffusion model, instead of the traditional 2-D convolution layers. Additionally, we introduce noise augmentation into the input and latent representations to improve the robustness and generalization of the generated results. Experiments in p-diff show that this augmentation is crucial for producing stable models with high performance.
Unlike generating model parameters trained on IID data in p-diff, our diffusion model is designed to fit model parameters from different data distributions in FL. For smaller models, we generate the entire model, while for larger models, we generate specific layers due to hardware constraints.
The complete details of the model are provided in the Appendix.

\section{Experiments} \label{sec:experiments}
\vspace{-0.15cm}

\subsection{Setup}
\paragraph{Datasets and Local Models.} We conduct image classification tasks and evaluate our method on three widely-used benchmark datasets: Fashion-MNIST \cite{fmnist}, EMNIST \cite{EMNIST17}, and CIFAR-10 \cite{cifar10/100}. A small CNN model for Fashion-MNIST and two larger CNN models for EMNIST and CIFAR-10 are constructed, respectively. Details of the datasets and model architectures are provided in the Appendix.

\paragraph{Data Partitioning.} For the heterogeneous data distribution setting, we adopt the approach proposed in \cite{Karimireddy20SCAFFOLD,Zhang21FedFOMO}, ensuring that all clients have equal data sizes. A portion of the data (s\%, default 20\%) is uniformly sampled from all classes, while the remaining (100 - s)\% is sampled from a set of dominant classes specific to each client. Clients are grouped based on their dominant classes, though this grouping is unknown to the server. Additionally, the size of the local training data is kept small, specifically at 600 samples per client, to emphasize the necessity of FL. The testing data on each client is drawn from the same distribution as the training data.

\paragraph{Compared Baselines.} We evaluate the performance of \texttt{pFedGPA} against the following baselines: Local-only, where each client trains its model locally without communication; FedAvg  \cite{mcmahan2017communication}and its locally fine-tuned variant (FedAvg-FT); pFedHN \cite{shamsian2021personalized}, which leverages a server-side hypernetwork; and other personalized FL methods, including FedPer \cite{arivazhagan2019fedper}, FedRep \cite{Collins21FedRep}, FedBABU \cite{OhKY22FedBABU}, and LG-FedAvg \cite{liang2020LGFedAvg}. Hyperparameters for all baselines were tuned to their optimal values based on the settings reported in the original papers or determined through grid search.

\paragraph{Training Details.}
All local models are trained using mini-batch SGD as the optimizer, with 2 epochs per round and a batch size of 50. The number of global communication rounds is set to 200 for Fashion-MNIST and EMNIST, and 300 for CIFAR-10. For Fashion-MNIST, the entire model is generated, while for EMNIST and CIFAR-10, only the final two fully connected layers are generated. We report the average test accuracy across clients. For \texttt{pFedGPA}, we trained the diffusion models using parameters collected from the last 20 rounds, generating new parameters in the final round.

\subsection{Experimental Results}

\paragraph{Performance Comparison.}
The results of average test accuracy across clients are reported in Table~\ref{tab:result}, where our \texttt{pFedGPA} is able to outperform other baselines in most cases. The improvement by our method increases as the learning task becomes harder. For example, our method improve the accuracy by $\sim$2\% with 100 clients on the CIFAR-10. The performance gaps between FedPer/FedRep and FedAvg-FT/FedBABU on various cases indicate that only sharing the feature extractor during the FL process could have inverse affect, as diverse local classifier head could make the feature learning sub-optimal. By contrast, our framework leverages the powerful generation ability of diffusion models to facilitate the personalized model aggregation.

\begin{figure}[t!]
	\centering
    \includegraphics[width=0.48\textwidth]
    {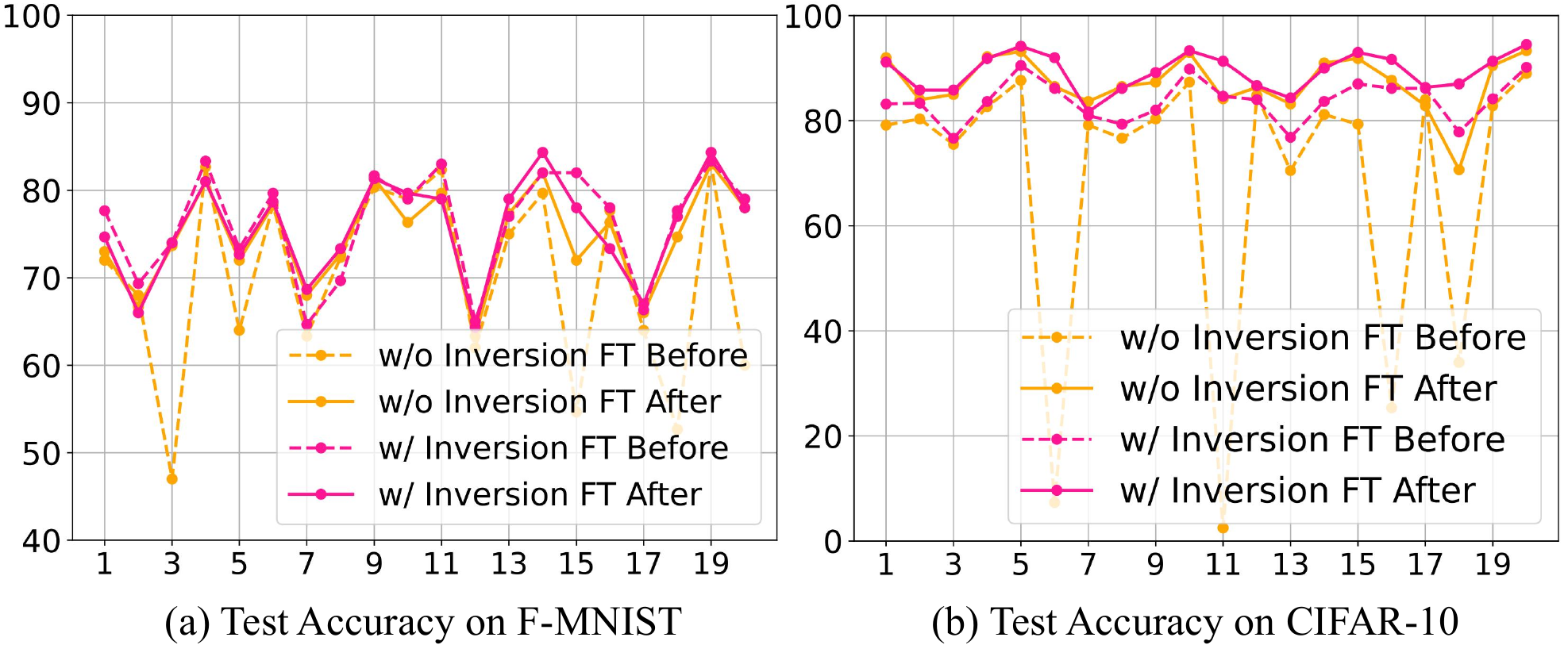}
        \vspace{-3ex}
	\caption{Comparison of test accuracies across 20 clients on Fashion-MNIST and CIFAR-10 using the \texttt{pFedGPA} method, before and after the fine-tuning (FT) operation, with and without parameter inversion. Dashed lines indicate results before FT, and solid lines indicate results after FT.}
	\vspace{-2pt}
	\label{result5}
\end{figure}

\paragraph{Parameter Initialization.}
The generative approach captures the distribution of model parameters across different clients and can provide parameter initialization for subsequent networks. The diffusion model trained within one FL network can subsequently serve as a pre-trained model for a newly initialized FL network, thus avoiding a cold start and accelerating the initialization process. To validate this idea, we employed a diffusion model trained within one FL network to generate parameters for a newly initialized network. We then compared the number of communication rounds required by each method to reach 95\% of its respective peak performance, measured as the average test accuracy. Experimental results, detailed in the appendix, demonstrate that our approach significantly reduces the number of communication rounds required to achieve high performance, compared to baseline methods.

		

\begin{table}[t]
    \centering
    \normalsize
    \resizebox{\columnwidth}{!}{
        \begin{tabular}{ c c c c c }
            \toprule[1pt]
            \multirow{2}{*}{Dataset} & \multicolumn{2}{c}{w/o inversion} & \multicolumn{2}{c}{w/ inversion} \\
            \cmidrule(lr){2-3}\cmidrule(lr){4-5}
            & before FT & after FT & before FT & after FT \\
            \midrule
            EMNIST (\%)  
            & 69.38 & 81.09 & 83.54 & \textbf{84.70}
            \\
            Fashion-MNIST (\%) 
            & 68.48 & 87.24 & 83.82 & \textbf{89.37}
            \\
            CIFAR-10 (\%) 
            & 68.9 & 74.77 & \textbf{76.28} & 75.72
            \\
            \bottomrule[1pt]
        \end{tabular}
    }
    \caption{Ablation study on the effect of the parameter inversion mechanism in \texttt{pFedGPA}. \textbf{before FT} and \textbf{after FT} refer to the accuracy before and after fine-tuning on local data in the final round. \textbf{w/o inversion} denotes without the inversion mechanism, and \textbf{w/ inversion} denotes with the inversion mechanism applied.}
    \label{tab:ablation}
    \vspace{-2ex}
\end{table}

\paragraph{Ablation Study.}
We conduct ablation studies to verify the efficacy of the inversion mechanism in \texttt{pFedGPA}. Specifically, we compare the performance of using inversion versus direct generation on the three datasets. We report the final round's average accuracy on 20 clients, both before and after fine-tuning. As demonstrated in Table \ref{tab:ablation}, the inversion mechanism significantly improve the average test accuracy before and after fine-tuning. Beyond the advantage of inversion in generating personalized parameters, we found that direct generation is less stable, often resulting in failed parameter generation.
We compared the test accuracies across 20 clients on Fashion-MNIST and CIFAR-10 using the pFedGPA method, evaluating the performance both before and after the fine-tuning (FT) operation, with and without parameter inversion. In the results shown in Fig. ~\ref{result5}, dashed lines represent the accuracies before fine-tuning, while solid lines show the accuracies after fine-tuning. In about 3 out of 20 cases, directly generated parameters had a pre-finetune accuracy below 60\%. While increasing the size of the diffusion model reduced this failure rate, inversion remained the more robust design, with failures occurring in only 1 out of 40 cases.

\subsection{Discussion}
\paragraph{Practical Considerations.} In our experiments, training a round of the diffusion model takes about an hour on a single Nvidia 4090 24GB GPU, with the entire FL process completing in four hours. However, the diffusion model's training time significantly exceeds the time between communication rounds, causing a bottleneck in the overall process. In real-world applications, it is expected that the server would have more hardware resources, thereby enabling the training of larger diffusion models, which can in turn generate larger local models. Additionally, employing larger batch sizes could further accelerate the training process. In contrast, the inference time remains relatively short, enabling the trained diffusion model to quickly generate parameters for clients.

\paragraph{Communication and Privacy Concerns.} The trained parameters of the diffusion model, which are much larger than those of the clients, are never transmitted. Consequently, we do not introduce any additional communication costs compared to other methods.
Moreover, in our framework, only the local model parameters are exchanged between clients and the server, without transmitting additional local feature statistics, thereby maintaining the same privacy levels as other methods that exchange only parameters.

\paragraph{Future Work.} Beyond the advantages demonstrated in our experiments, generative aggregation methods have the potential to be integrated with other FL approaches, such as prototype-enhanced algorithms \cite{tan2022fedproto,xu2023personalized}. These methods can complement each other effectively: generative models learn the distribution of model parameters, while prototype learning focuses on the distribution of data representations. This integration could enable the development of novel inversion strategies that may lead to the generation of more accurate personalized parameters in the future.

\section{Conclusion}\label{sec:conclusion}
In this paper, we introduce a novel diffusion-based parameter aggregation method for personalized FL. In our framework,
the server deploys a diffusion model to consolidate the uploaded
parameters and generates personalized parameters for each client with the global guidance by using a newly developed inversion mechanism.
Experimental results on three datasets verify the effectiveness and further advantages of our method. 

\section{Acknowledgments}
This work is supported in part by the Natural Science Foundation of China (Grant 62371270)  and  Shenzhen Key Laboratory of Ubiquitous Data Enabling (No.ZDSYS20220527171406015).

\bibliography{aaai25}

\begin{thebibliography}{56}
\providecommand{\natexlab}[1]{#1}

\bibitem[{Amari(1993)}]{amari1993backpropagation}
Amari, S.-i. 1993.
\newblock Backpropagation and stochastic gradient descent method.
\newblock \emph{Neurocomputing}, 5(4-5): 185--196.

\bibitem[{Arivazhagan et~al.(2019)Arivazhagan, Aggarwal, Singh, and Choudhary}]{arivazhagan2019fedper}
Arivazhagan, M.~G.; Aggarwal, V.; Singh, A.~K.; and Choudhary, S. 2019.
\newblock Federated learning with personalization layers.
\newblock \emph{arXiv preprint arXiv:1912.00818}.

\bibitem[{Cohen et~al.(2017)Cohen, Afshar, Tapson, and van Schaik}]{EMNIST17}
Cohen, G.; Afshar, S.; Tapson, J.; and van Schaik, A. 2017.
\newblock {EMNIST:} Extending {MNIST} to handwritten letters.
\newblock In \emph{2017 International Joint Conference on Neural Networks, {IJCNN} 2017, Anchorage, AK, USA, May 14-19, 2017}, 2921--2926. {IEEE}.

\bibitem[{Collins et~al.(2021)Collins, Hassani, Mokhtari, and Shakkottai}]{Collins21FedRep}
Collins, L.; Hassani, H.; Mokhtari, A.; and Shakkottai, S. 2021.
\newblock Exploiting Shared Representations for Personalized Federated Learning.
\newblock In \emph{Proceedings of the 38th International Conference on Machine Learning, {ICML} 2021, 18-24 July 2021, Virtual Event}, volume 139 of \emph{Proceedings of Machine Learning Research}, 2089--2099. {PMLR}.

\bibitem[{Dhariwal and Nichol(2021)}]{dhariwal2021diffusion}
Dhariwal, P.; and Nichol, A. 2021.
\newblock Diffusion models beat gans on image synthesis.
\newblock \emph{Advances in neural information processing systems}, 34: 8780--8794.

\bibitem[{Erko{\c{c}} et~al.(2023)Erko{\c{c}}, Ma, Shan, Nie{\ss}ner, and Dai}]{erkocc2023hyperdiffusion}
Erko{\c{c}}, Z.; Ma, F.; Shan, Q.; Nie{\ss}ner, M.; and Dai, A. 2023.
\newblock Hyperdiffusion: Generating implicit neural fields with weight-space diffusion.
\newblock In \emph{Proceedings of the IEEE/CVF international conference on computer vision}, 14300--14310.

\bibitem[{Fallah, Mokhtari, and Ozdaglar(2020)}]{fallah2020personalized}
Fallah, A.; Mokhtari, A.; and Ozdaglar, A. 2020.
\newblock Personalized federated learning with theoretical guarantees: A model-agnostic meta-learning approach.
\newblock \emph{Advances in neural information processing systems}, 33: 3557--3568.

\bibitem[{Finn, Abbeel, and Levine(2017)}]{finn2017model}
Finn, C.; Abbeel, P.; and Levine, S. 2017.
\newblock Model-agnostic meta-learning for fast adaptation of deep networks.
\newblock In \emph{International conference on machine learning}, 1126--1135. PMLR.

\bibitem[{Hanzely and Richt{\'a}rik(2020)}]{hanzely2020federated}
Hanzely, F.; and Richt{\'a}rik, P. 2020.
\newblock Federated learning of a mixture of global and local models.
\newblock \emph{arXiv preprint arXiv:2002.05516}.

\bibitem[{Ho, Jain, and Abbeel(2020)}]{ho2020denoising}
Ho, J.; Jain, A.; and Abbeel, P. 2020.
\newblock Denoising diffusion probabilistic models.
\newblock \emph{Advances in neural information processing systems}, 33: 6840--6851.

\bibitem[{Ho and Salimans(2022)}]{ho2022classifier}
Ho, J.; and Salimans, T. 2022.
\newblock Classifier-free diffusion guidance.
\newblock \emph{arXiv preprint arXiv:2207.12598}.

\bibitem[{Kairouz et~al.(2021)Kairouz, McMahan, Avent, Bellet, Bennis, Bhagoji, Bonawitz, Charles, Cormode, Cummings et~al.}]{kairouz2021advances}
Kairouz, P.; McMahan, H.~B.; Avent, B.; Bellet, A.; Bennis, M.; Bhagoji, A.~N.; Bonawitz, K.; Charles, Z.; Cormode, G.; Cummings, R.; et~al. 2021.
\newblock Advances and open problems in federated learning.
\newblock \emph{Foundations and trends{\textregistered} in machine learning}, 14(1--2): 1--210.

\bibitem[{Karimireddy et~al.(2020{\natexlab{a}})Karimireddy, Kale, Mohri, Reddi, Stich, and Suresh}]{karimireddy2020scaffold}
Karimireddy, S.~P.; Kale, S.; Mohri, M.; Reddi, S.; Stich, S.; and Suresh, A.~T. 2020{\natexlab{a}}.
\newblock Scaffold: Stochastic controlled averaging for federated learning.
\newblock In \emph{International conference on machine learning}, 5132--5143. PMLR.

\bibitem[{Karimireddy et~al.(2020{\natexlab{b}})Karimireddy, Kale, Mohri, Reddi, Stich, and Suresh}]{Karimireddy20SCAFFOLD}
Karimireddy, S.~P.; Kale, S.; Mohri, M.; Reddi, S.~J.; Stich, S.~U.; and Suresh, A.~T. 2020{\natexlab{b}}.
\newblock {SCAFFOLD:} Stochastic Controlled Averaging for Federated Learning.
\newblock In \emph{Proceedings of the 37th International Conference on Machine Learning, {ICML} 2020, 13-18 July 2020, Virtual Event}, volume 119 of \emph{Proceedings of Machine Learning Research}, 5132--5143. {PMLR}.

\bibitem[{Kawar et~al.(2023)Kawar, Zada, Lang, Tov, Chang, Dekel, Mosseri, and Irani}]{kawar2023imagic}
Kawar, B.; Zada, S.; Lang, O.; Tov, O.; Chang, H.; Dekel, T.; Mosseri, I.; and Irani, M. 2023.
\newblock Imagic: Text-based real image editing with diffusion models.
\newblock In \emph{Proceedings of the IEEE/CVF Conference on Computer Vision and Pattern Recognition}, 6007--6017.

\bibitem[{Krizhevsky and Hinton(2009)}]{cifar10/100}
Krizhevsky, A.; and Hinton, G. 2009.
\newblock Learning Multiple Layers of Features from Tiny Images.
\newblock \emph{University of Toronto}.

\bibitem[{Kumari et~al.(2023)Kumari, Zhang, Zhang, Shechtman, and Zhu}]{kumari2023multi}
Kumari, N.; Zhang, B.; Zhang, R.; Shechtman, E.; and Zhu, J.-Y. 2023.
\newblock Multi-concept customization of text-to-image diffusion.
\newblock In \emph{Proceedings of the IEEE/CVF Conference on Computer Vision and Pattern Recognition}, 1931--1941.

\bibitem[{Li, He, and Song(2021)}]{li2021model}
Li, Q.; He, B.; and Song, D. 2021.
\newblock Model-contrastive federated learning.
\newblock In \emph{Proceedings of the IEEE/CVF conference on computer vision and pattern recognition}, 10713--10722.

\bibitem[{Li et~al.(2020)Li, Sahu, Zaheer, Sanjabi, Talwalkar, and Smith}]{li2020federated}
Li, T.; Sahu, A.~K.; Zaheer, M.; Sanjabi, M.; Talwalkar, A.; and Smith, V. 2020.
\newblock Federated optimization in heterogeneous networks.
\newblock \emph{Proceedings of Machine learning and systems}, 2: 429--450.

\bibitem[{Li et~al.(2019)Li, Milletar{\`\i}, Xu, Rieke, Hancox, Zhu, Baust, Cheng, Ourselin, Cardoso et~al.}]{li2019privacy}
Li, W.; Milletar{\`\i}, F.; Xu, D.; Rieke, N.; Hancox, J.; Zhu, W.; Baust, M.; Cheng, Y.; Ourselin, S.; Cardoso, M.~J.; et~al. 2019.
\newblock Privacy-preserving federated brain tumour segmentation.
\newblock In \emph{Machine Learning in Medical Imaging: 10th International Workshop, MLMI 2019, Held in Conjunction with MICCAI 2019, Shenzhen, China, October 13, 2019, Proceedings 10}, 133--141. Springer.

\bibitem[{Liang et~al.(2020)Liang, Liu, Ziyin, Allen, Auerbach, Brent, Salakhutdinov, and Morency}]{liang2020LGFedAvg}
Liang, P.~P.; Liu, T.; Ziyin, L.; Allen, N.~B.; Auerbach, R.~P.; Brent, D.; Salakhutdinov, R.; and Morency, L.-P. 2020.
\newblock Think locally, act globally: Federated learning with local and global representations.
\newblock \emph{arXiv preprint arXiv:2001.01523}.

\bibitem[{Lugmayr et~al.(2022)Lugmayr, Danelljan, Romero, Yu, Timofte, and Van~Gool}]{lugmayr2022repaint}
Lugmayr, A.; Danelljan, M.; Romero, A.; Yu, F.; Timofte, R.; and Van~Gool, L. 2022.
\newblock Repaint: Inpainting using denoising diffusion probabilistic models.
\newblock In \emph{Proceedings of the IEEE/CVF conference on computer vision and pattern recognition}, 11461--11471.

\bibitem[{McMahan et~al.(2017{\natexlab{a}})McMahan, Moore, Ramage, Hampson, and y~Arcas}]{mcmahan2017communication}
McMahan, B.; Moore, E.; Ramage, D.; Hampson, S.; and y~Arcas, B.~A. 2017{\natexlab{a}}.
\newblock Communication-efficient learning of deep networks from decentralized data.
\newblock In \emph{Artificial intelligence and statistics}, 1273--1282. PMLR.

\bibitem[{McMahan et~al.(2017{\natexlab{b}})McMahan, Moore, Ramage et~al.}]{mcmahan2016communication}
McMahan, H.~B.; Moore, E.; Ramage, D.; et~al. 2017{\natexlab{b}}.
\newblock Communication-efficient learning of deep networks from decentralized data.
\newblock \emph{AISTATS}.

\bibitem[{Meng et~al.(2021)Meng, He, Song, Song, Wu, Zhu, and Ermon}]{meng2021sdedit}
Meng, C.; He, Y.; Song, Y.; Song, J.; Wu, J.; Zhu, J.-Y.; and Ermon, S. 2021.
\newblock Sdedit: Guided image synthesis and editing with stochastic differential equations.
\newblock \emph{arXiv preprint arXiv:2108.01073}.

\bibitem[{Nguyen et~al.(2022)Nguyen, Pham, Pathirana, Ding, Seneviratne, Lin, Dobre, and Hwang}]{nguyen2022health}
Nguyen, D.~C.; Pham, Q.-V.; Pathirana, P.~N.; Ding, M.; Seneviratne, A.; Lin, Z.; Dobre, O.; and Hwang, W.-J. 2022.
\newblock Federated learning for smart healthcare: A survey.
\newblock \emph{ACM Computing Surveys (Csur)}, 55(3): 1--37.

\bibitem[{Nichol, Achiam, and Schulman(2018)}]{nichol2018first}
Nichol, A.; Achiam, J.; and Schulman, J. 2018.
\newblock On first-order meta-learning algorithms.
\newblock \emph{arXiv preprint arXiv:1803.02999}.

\bibitem[{Oh, Kim, and Yun(2022)}]{OhKY22FedBABU}
Oh, J.; Kim, S.; and Yun, S. 2022.
\newblock FedBABU: Toward Enhanced Representation for Federated Image Classification.
\newblock In \emph{The Tenth International Conference on Learning Representations, {ICLR} 2022, Virtual Event, April 25-29, 2022}.

\bibitem[{Peebles et~al.(2022)Peebles, Radosavovic, Brooks, Efros, and Malik}]{peebles2022learning}
Peebles, W.; Radosavovic, I.; Brooks, T.; Efros, A.~A.; and Malik, J. 2022.
\newblock Learning to learn with generative models of neural network checkpoints.
\newblock \emph{arXiv preprint arXiv:2209.12892}.

\bibitem[{Ramesh et~al.(2022)Ramesh, Dhariwal, Nichol, Chu, and Chen}]{ramesh2022hierarchical}
Ramesh, A.; Dhariwal, P.; Nichol, A.; Chu, C.; and Chen, M. 2022.
\newblock Hierarchical text-conditional image generation with clip latents.
\newblock \emph{arXiv preprint arXiv:2204.06125}, 1(2): 3.

\bibitem[{Rombach et~al.(2022)Rombach, Blattmann, Lorenz, Esser, and Ommer}]{rombach2022high}
Rombach, R.; Blattmann, A.; Lorenz, D.; Esser, P.; and Ommer, B. 2022.
\newblock High-resolution image synthesis with latent diffusion models.
\newblock In \emph{Proceedings of the IEEE/CVF conference on computer vision and pattern recognition}, 10684--10695.

\bibitem[{Saharia et~al.(2022)Saharia, Chan, Saxena, Li, Whang, Denton, Ghasemipour, Gontijo~Lopes, Karagol~Ayan, Salimans et~al.}]{saharia2022photorealistic}
Saharia, C.; Chan, W.; Saxena, S.; Li, L.; Whang, J.; Denton, E.~L.; Ghasemipour, K.; Gontijo~Lopes, R.; Karagol~Ayan, B.; Salimans, T.; et~al. 2022.
\newblock Photorealistic text-to-image diffusion models with deep language understanding.
\newblock \emph{Advances in neural information processing systems}, 35: 36479--36494.

\bibitem[{Sattler, M{\"u}ller, and Samek(2020)}]{sattler2020clustered}
Sattler, F.; M{\"u}ller, K.-R.; and Samek, W. 2020.
\newblock Clustered federated learning: Model-agnostic distributed multitask optimization under privacy constraints.
\newblock \emph{IEEE transactions on neural networks and learning systems}, 32(8): 3710--3722.

\bibitem[{Shamsian et~al.(2021)Shamsian, Navon, Fetaya, and Chechik}]{shamsian2021personalized}
Shamsian, A.; Navon, A.; Fetaya, E.; and Chechik, G. 2021.
\newblock Personalized federated learning using hypernetworks.
\newblock In \emph{International Conference on Machine Learning}, 9489--9502. PMLR.

\bibitem[{Smith et~al.(2017)Smith, Chiang, Sanjabi, and Talwalkar}]{smith2017federated}
Smith, V.; Chiang, C.-K.; Sanjabi, M.; and Talwalkar, A.~S. 2017.
\newblock Federated multi-task learning.
\newblock \emph{Advances in neural information processing systems}, 30.

\bibitem[{Sohl-Dickstein et~al.(2015)Sohl-Dickstein, Weiss, Maheswaranathan, and Ganguli}]{sohl2015deep}
Sohl-Dickstein, J.; Weiss, E.; Maheswaranathan, N.; and Ganguli, S. 2015.
\newblock Deep unsupervised learning using nonequilibrium thermodynamics.
\newblock In \emph{International conference on machine learning}, 2256--2265. PMLR.

\bibitem[{Song and Ermon(2019)}]{song2019generative}
Song, Y.; and Ermon, S. 2019.
\newblock Generative modeling by estimating gradients of the data distribution.
\newblock \emph{Advances in neural information processing systems}, 32.

\bibitem[{Song et~al.(2020)Song, Sohl-Dickstein, Kingma, Kumar, Ermon, and Poole}]{song2020score}
Song, Y.; Sohl-Dickstein, J.; Kingma, D.~P.; Kumar, A.; Ermon, S.; and Poole, B. 2020.
\newblock Score-based generative modeling through stochastic differential equations.
\newblock \emph{arXiv preprint arXiv:2011.13456}.

\bibitem[{Tan et~al.(2022)Tan, Long, Liu, Zhou, Lu, Jiang, and Zhang}]{tan2022fedproto}
Tan, Y.; Long, G.; Liu, L.; Zhou, T.; Lu, Q.; Jiang, J.; and Zhang, C. 2022.
\newblock Fedproto: Federated prototype learning across heterogeneous clients.
\newblock In \emph{Proceedings of the AAAI Conference on Artificial Intelligence}, volume~36, 8432--8440.

\bibitem[{Wang et~al.(2024)Wang, Xu, Zhou, Zang, Darrell, Liu, and You}]{wang2024neural}
Wang, K.; Xu, Z.; Zhou, Y.; Zang, Z.; Darrell, T.; Liu, Z.; and You, Y. 2024.
\newblock Neural network diffusion.
\newblock \emph{arXiv preprint arXiv:2402.13144}.

\bibitem[{Wilson and Izmailov(2020)}]{wilson2020bayesian}
Wilson, A.~G.; and Izmailov, P. 2020.
\newblock Bayesian deep learning and a probabilistic perspective of generalization.
\newblock \emph{Advances in neural information processing systems}, 33: 4697--4708.

\bibitem[{Wu and De~la Torre(2023)}]{wu2023latent}
Wu, C.~H.; and De~la Torre, F. 2023.
\newblock A latent space of stochastic diffusion models for zero-shot image editing and guidance.
\newblock In \emph{Proceedings of the IEEE/CVF International Conference on Computer Vision}, 7378--7387.

\bibitem[{Xiao, Rasul, and Vollgraf(2017)}]{fmnist}
Xiao, H.; Rasul, K.; and Vollgraf, R. 2017.
\newblock Fashion-MNIST: a Novel Image Dataset for Benchmarking Machine Learning Algorithms.
\newblock \emph{CoRR}, abs/1708.07747.

\bibitem[{Xu, Tong, and Huang(2023)}]{xu2023personalized}
Xu, J.; Tong, X.; and Huang, S.-L. 2023.
\newblock Personalized Federated Learning with Feature Alignment and Classifier Collaboration.
\newblock In \emph{The Eleventh International Conference on Learning Representations}.

\bibitem[{Xu, Yan, and Huang(2022)}]{Xu22FedPer++}
Xu, J.; Yan, Y.; and Huang, S. 2022.
\newblock FedPer++: Toward Improved Personalized Federated Learning on Heterogeneous and Imbalanced Data.
\newblock In \emph{{IJCNN}}.

\bibitem[{Xu et~al.(2022)Xu, Yu, Song, Shi, Ermon, and Tang}]{xu2022geodiff}
Xu, M.; Yu, L.; Song, Y.; Shi, C.; Ermon, S.; and Tang, J. 2022.
\newblock Geodiff: A geometric diffusion model for molecular conformation generation.
\newblock \emph{arXiv preprint arXiv:2203.02923}.

\bibitem[{Yang et~al.(2023)Yang, Xu, Ding, and Liu}]{Yang23FedCap}
Yang, M.; Xu, J.; Ding, W.; and Liu, Y. 2023.
\newblock Federated Capsule Graph Neural Network with Personalization.
\newblock In \emph{Adjunct Proceedings of the 2022 ACM International Joint Conference on Pervasive and Ubiquitous Computing and the 2022 ACM International Symposium on Wearable Computers}, UbiComp/ISWC '22 Adjunct.

\bibitem[{Yang et~al.(2019)Yang, Liu, Chen, and Tong}]{yang2019federated}
Yang, Q.; Liu, Y.; Chen, T.; and Tong, Y. 2019.
\newblock Federated machine learning: Concept and applications.
\newblock \emph{ACM Transactions on Intelligent Systems and Technology (TIST)}, 10(2): 1--19.

\bibitem[{Ye et~al.(2023)Ye, Ni, Wu, Chen, and Wang}]{ye23fedgraph}
Ye, R.; Ni, Z.; Wu, F.; Chen, S.; and Wang, Y. 2023.
\newblock Personalized Federated Learning with Inferred Collaboration Graphs.
\newblock In \emph{Proceedings of the 40th International Conference on Machine Learning}, volume 202 of \emph{Proceedings of Machine Learning Research}, 39801--39817. PMLR.

\bibitem[{Zhang et~al.(2024)Zhang, Luo, Yu, Li, Lin, Ye, and Zhang}]{zhang2024metadiff}
Zhang, B.; Luo, C.; Yu, D.; Li, X.; Lin, H.; Ye, Y.; and Zhang, B. 2024.
\newblock Metadiff: Meta-learning with conditional diffusion for few-shot learning.
\newblock In \emph{Proceedings of the AAAI Conference on Artificial Intelligence}, volume~38, 16687--16695.

\bibitem[{Zhang et~al.(2023)Zhang, Hua, Wang, Song, Xue, Ma, and Guan}]{zhang2023fedala}
Zhang, J.; Hua, Y.; Wang, H.; Song, T.; Xue, Z.; Ma, R.; and Guan, H. 2023.
\newblock Fedala: Adaptive local aggregation for personalized federated learning.
\newblock In \emph{Proceedings of the AAAI Conference on Artificial Intelligence}, volume~37, 11237--11244.

\bibitem[{Zhang et~al.(2021)Zhang, Sapra, Fidler, Yeung, and Alvarez}]{Zhang21FedFOMO}
Zhang, M.; Sapra, K.; Fidler, S.; Yeung, S.; and Alvarez, J.~M. 2021.
\newblock Personalized Federated Learning with First Order Model Optimization.
\newblock In \emph{9th International Conference on Learning Representations, {ICLR} 2021, Virtual Event, Austria, May 3-7, 2021}. OpenReview.net.

\bibitem[{Zhao et~al.(2018)Zhao, Li, Lai, Suda, Civin, and Chandra}]{zhao2018federated}
Zhao, Y.; Li, M.; Lai, L.; Suda, N.; Civin, D.; and Chandra, V. 2018.
\newblock Federated learning with non-iid data.
\newblock \emph{arXiv:1806.00582}.

\bibitem[{Zhmoginov, Sandler, and Vladymyrov(2022)}]{zhmoginov2022hypertransformer}
Zhmoginov, A.; Sandler, M.; and Vladymyrov, M. 2022.
\newblock Hypertransformer: Model generation for supervised and semi-supervised few-shot learning.
\newblock In \emph{International Conference on Machine Learning}, 27075--27098. PMLR.

\bibitem[{Zhou et~al.(2022)Zhou, Qixuan, Luo, Zhang, and Xu}]{zhou2022towards}
Zhou, H.; Qixuan, Z.; Luo, T.; Zhang, Y.; and Xu, Z.-Q. 2022.
\newblock Towards understanding the condensation of neural networks at initial training.
\newblock \emph{Advances in Neural Information Processing Systems}, 35: 2184--2196.

\bibitem[{Zhou et~al.(2023)Zhou, Zhou, Li, and Xu}]{zhou2023understanding}
Zhou, Z.; Zhou, H.; Li, Y.; and Xu, Z.-Q.~J. 2023.
\newblock Understanding the initial condensation of convolutional neural networks.
\newblock \emph{arXiv preprint arXiv:2305.09947}.

\end{thebibliography}

\end{document}